\RequirePackage{amsthm}
\documentclass[sn-mathphys-num]{sn-jnl}

\usepackage{amsmath,amssymb,amsfonts}
\usepackage{bm}
\usepackage{mathrsfs}%
\usepackage{graphicx}
\usepackage{textcomp}
\usepackage{xcolor}
\usepackage{hyperref}
\usepackage{multirow}
\usepackage{url}
\usepackage{algorithm}
\usepackage{algpseudocode}

\newtheorem{theorem}{Theorem}
\newtheorem{lemma}{Lemma}

\usepackage{mathtools}
\DeclareMathOperator*{\argmax}{arg\,max}

\usepackage[title]{appendix}
\usepackage{subcaption}
\usepackage{booktabs}%
\theoremstyle{thmstyleone}%
\theoremstyle{thmstyletwo}%
\theoremstyle{thmstylethree}%
\usepackage{lipsum}

\begin{document}

\title[Article Title]{Directly Classification for missing Electronic Healthcare Records Data with Explainable Artificial Intelligence}


\title{Directly Handling Missing Data in Linear Discriminant Analysis for Enhancing Classification Accuracy and Interpretability}

\author{Tuan L. Vo}

\author{Uyen Dang}

\author{Thu Nguyen}





\abstract{As the adoption of Artificial Intelligence (AI) models expands into critical real-world applications, ensuring the explainability of these models becomes paramount, particularly in sensitive fields such as medicine and finance. Linear Discriminant Analysis (LDA) remains a popular choice for classification due to its interpretable nature, derived from its capacity to model class distributions and enhance class separation through linear combinations of features. However, real-world datasets often suffer from incomplete data, posing substantial challenges for both classification accuracy and model interpretability. In this paper, we introduce a novel and robust classification method, \textbf{termed Weighted missing Linear Discriminant Analysis (WLDA)}, which extends LDA to handle datasets with missing values without the need for imputation. Our approach innovatively incorporates a weight matrix that penalizes missing entries, thereby refining parameter estimation directly on incomplete data. This methodology not only preserves the interpretability of LDA but also significantly enhances classification performance in scenarios plagued by missing data. We conduct an in-depth theoretical analysis to establish the properties of WLDA and thoroughly evaluate its explainability. Experimental results across various datasets demonstrate that WLDA consistently outperforms traditional methods, especially in challenging environments where missing values are prevalent in both training and test datasets. This advancement provides a critical tool for improving classification accuracy and maintaining model transparency in the face of incomplete data.}

\maketitle
\section{Introduction}\label{sec:introduction}
 
Despite achieving powerful performance in many real-world scenarios, the lack of explainability in decisions made by ``black box" AI models is a major barrier to their adoption in high-stakes fields across various domains~\cite{samek2019towards,nguyen2023multimedia}. For example, in the medical field, transparency is necessary for doctors to use AI models in practical scenarios. Explainable AI (XAI) aims to address this issue by creating explanations for the decision-making processes of machine learning models, allowing users to understand and validate AI actions, thus facilitating greater trust and wider application of AI technologies~\cite{minh2022explainable,dovsilovic2018explainable}. Additionally, XAI helps identify and rectify biases and errors in AI models, enhancing their reliability and fairness. Ultimately, XAI bridges the gap between complex AI algorithms and human comprehension, fostering more ethical, reliable, and user-friendly AI applications.

Linear Discriminant Analysis (LDA) is a classical technique that offers several advantages compared to more recently introduced techniques such as XGBoost, Neural Networks, and Support Vector Machines (SVM): lower computational demand, making it well-suited for large datasets; less prone to overfitting due to its limited number of adjustable parameters and reliance on linear assumptions; does not require extensive hyperparameter tuning;  robust against noisy data and outliers. Importantly, it is a widely recognized interpretable model that models class distributions and identifies the linear combination of features that maximizes class separation~\cite{tharwat2016linear}. However, missing data is a common problem in practice, and the presence of missing data can seriously affect the parameter estimation and classification, as well as the interpretation process of LDA.
To address issues presented by incomplete data, the most prevalent strategy involves data imputation - a process where missing data points are filled in using various estimation techniques. However, the limitation of this approach is that it can introduce bias and inaccuracies if the imputed values do not accurately reflect the underlying data distribution. Additionally, imputation methods often require assumptions about the data, which may not hold true in all cases, leading to suboptimal model performance. 

The above analysis motivates us to propose \textbf{\textit{Weighted missing Linear Discriminant Analysis (WLDA)}} to address the challenges with LDA under missing data. More specifically, we want 
to address the dual challenges of handling missing data and ensuring model explainability in machine learning applications. 
The fundamental idea of the method is to use the DPER algorithm \cite{NGUYEN2022108082} for estimating the parameters, and the idea that when a value is missing, its contribution to classification should be zero. To do that, they utilize a weight matrix to penalize the contribution of a missing value. As will be illustrated experimentally, the proposed WLDA algorithm not only enhances the robustness and accuracy of models in the presence of missing data but also ensures that the decision-making process is transparent and comprehensible. This dual focus on data integrity and explainability is crucial for fostering trust, reliability, and broader acceptance of AI technologies in critical applications.
Also, the proposed technique can be explained throughout from pre-modeling to post-modeling. In other words, we aim to adjust the LDA model to directly learn from missing data and give an explanation for this model. To achieve this, we use a weighted missing matrix to evaluate the contribution of each feature when it contains missing values. 
In summary, our contribution can be summarized as follows:
(1) We introduce a novel algorithm, namely WLDA, capable of handling missing values in both training and test sets and treating each feature fairly without imputation;
 (ii) We theoretically analyze WLDA properties;
 (iii) We examine the explainability of the proposed algorithm in various aspects;
 (iv) We conduct various experiments to demonstrate that our algorithm achieves superior performance compared to other methods under comparison;
(v) We analyze the potential direction for future works.

The rest of this paper is organized as follows: Section~\ref{sec:related_works} provides an overview of related works on  XAI, LDA, and missing data. Next, we provide some preliminaries on LDA and its decision boundaries in Section~\ref{sec:preliminary}. Then, in Section~\ref{sec:methodologies}, we provide methodologies and then present the WLDA algorithm in~\ref{sec:LDA_missing}, the description of this algorithm can be found in~\ref{description}, and theoretical analysis for the WLDA algorithm is presented in~\ref{sec:theoretical_analysis}. Next, we detail the explanation strategies for our model in Section~\ref{sec:Explanation}. After that, in Section~\ref{sec:experiments}, we illustrate the efficiency of our algorithm via experiments and analyze the results. Finally, the paper ends with conclusions and our future works in Section~\ref{sec:conclusion}.

\section{Related works}\label{sec:related_works}

Explainability in machine learning models can be categorized into pre-modeling explainability, interpretable models, and post-modeling explainability~\cite{minh2022explainable}. 
In terms of explainability, LDA offers pre-modeling explainability~\cite{minh2022explainable} as it 
provides insights by interpreting class distributions and identifying the linear combination of features that maximizes class separation \cite{minh2022explainable}. LDA's decision boundaries and coefficients further enhance model explainability by making it clear how each feature contributes to classification decisions \cite{kuhn2013discriminant}. 

Also, LDA enhances model explainability via its decision boundaries and coefficients~\cite{kuhn2013discriminant}. It assumes that different classes generate data based on Gaussian distributions with the same covariance matrix but different means, resulting in linear decision boundaries between classes. These boundaries, determined by a linear combination of features, make it straightforward to understand each feature's contribution to classification decisions. Scatter plots with decision boundaries can visually demonstrate how the model separates different classes based on features, showing the degree of class separation and the validity of the linear assumption. The coefficients of the linear discriminants, akin to weights in linear regression, indicate each feature's importance and contribution to the decision-making process. By analyzing these coefficients, one can discern which features are most influential in distinguishing between classes.

In the post-modeling explainability of LDA, one can use 
Shapley values \cite{lundberg2017unified}, an idea derived from cooperative game theory.
Shapley values can pinpoint the significance of each feature value, thereby improving the transparency and interpretability of the model's results. Numerous studies have underscored the effectiveness of Shapley values in interpreting machine learning models. For instance, Lundberg and Lee introduced SHAP (SHapley Additive exPlanations)~\cite{lundberg2017unified}, a method based on Shapley values, which has proven to provide consistent and reliable explanations across various models. Likewise, Molnar~\cite{molnar2020interpretable} stressed the importance of model interpretability and highlighted the role of methods like Shapley values in making complex machine learning models more understandable. Integrating Shapley values with LDA paves the way for new approaches in interpreting topic models, allowing researchers and practitioners to make more informed and well-supported decisions based on LDA results.
In a different application, Cohen et al. demonstrated that Shapley-based explainable AI significantly improves clustering quality in fault diagnosis and prognosis applications, leading to information-dense and meaningful clusters~\cite{cohen2023shapley}. PWSHAP was introduced by Ter-Minassian~\cite{ter2023pwshap}, which is a framework that combines on-manifold Shapley values with path-wise effects to assess the targeted effect of binary variables in complex outcome models. Moreover, Kuroki et al. proposed BSED, a Baseline Shapley-based Explainable Detector for object detection, which extends the Shapley value to enhance the interpretability of predictions~\cite{kuroki2024bsed}.

The most common way to address missing values is through imputation methods. Techniques such as matrix decomposition or matrix completion, including Polynomial Matrix Completion~\cite{fan2020polynomial}, ALS~\cite{hastie2015matrix}, and Nuclear Norm Minimization~\cite{candes2009exact}, are employed to handle continuous data, making it complete for regular data analysis. Other methods involve regression or clustering, like CBRL and CBRC~\cite{m2020cbrl}, which use Bayesian Ridge Regression and cluster-based local least square methods~\cite{keerin2013improvement}. For large datasets, deep learning imputation techniques have shown good performance~\cite{choudhury2019imputation, garg2018dl, mohan2021graphical}. Different imputation approaches may yield varying values for each missing entry. Therefore, modeling the uncertainty for each missing entry is also crucial. Bayesian or multiple imputation techniques, such as Bayesian principal component analysis-based imputation~\cite{audigier2016multiple} and multiple imputations using Deep Denoising Autoencoders~\cite{gondara2018mida}, are often preferred. Tree-based techniques, like missForest~\cite{stekhoven2012missforest}, the DMI algorithm~\cite{rahman2013missing}, and DIFC~\cite{nikfalazar2020missing}, can naturally handle missing data through prediction. Recent methods that handle mixed data include SICE~\cite{khan2020sice}, FEMI~\cite{rahman2016missing}, and HCMM-LD~\cite{murray2016multiple}.

Recent studies combine imputation with the target task or adapt models to learn directly from missing data. For example, Dinh et al.~\cite{dinh2021clustering} integrate imputation and clustering into a single process. Ghazi et al.~\cite{ghazi2018robust} model disease progression using an LSTM architecture that handles missing data during both input and target phases, utilizing batch gradient descent with back-propagation. While these approaches may offer speed advantages and reduce storage costs, their complexity, and lack of generalizability across different datasets limit their application~\cite{nguyen4260235pmf}. Conversely, traditional imputation methods make data complete, facilitating generalization across datasets and enabling analysis with various techniques. However, most imputation methods lack explainability. Hans et al.~\cite{hans2023explainable} present an explainable imputation method based on association constraints in data, where explanations stem from the constraints used. Yet, this method requires known relationships or restrictions and has only been tested on small datasets, raising questions about its scalability. The DIMV~\cite{vu2023conditional} imputation method estimates the conditional distribution of a missing entry based on observed ones through direct parameter estimation by relying on direct parameter estimation~\cite{NGUYEN20211,NGUYEN2022108082}.


\vspace{-1mm}

\section{Preliminary: Linear Discriminant Analysis}\label{sec:preliminary}
We first have some notations which are used throughout this paper. Suppose that we have a data $\mathcal{D}=(\mathbf{X}|\mathbf{y})$ where $\mathbf{X}$ is size of $p \times n$ ($p$ features and $n$ instances), i.e.,
\begin{align*}
    \mathbf{X} = \begin{pmatrix}
        x_{11} & x_{12} & x_{13} & \dots & x_{1n} \\
        x_{11} & x_{12} & x_{23} & \dots & x_{1n} \\
        \vdots & \vdots & \vdots & \ddots & \vdots\\
        x_{p1} & x_{p2} & x_{3n} & \dots & x_{pn}
    \end{pmatrix},
\end{align*}
and $\mathbf{y}$ is the label. Also, suppose that the data is from $G$ classes, and each observation in $\mathbf{X}$ follows a multivariate normal distribution with mean $\boldsymbol{\mu}_{p \times 1}$ and covariance matrix $\boldsymbol{\Sigma}_{p \times p}$, i.e., $\mathbf{X} \sim \mathcal{N}(\boldsymbol{\mu},\boldsymbol{\Sigma})$. For each class $g$ where $g\in \{1,2,\dots,G\}$, denote $\mathbf{X}^{(g)}$ as the data from the $g^{th}$ class and $n_g$ is the number of observations in $\mathbf{X}^{(g)}$. Also, suppose that the observations in $\mathbf{X}^{(g)}$ follow a multivariate normal distribution $\mathcal{N}(\boldsymbol{\mu}^{(g)},\boldsymbol{\Sigma})$, i.e., we are assuming the observations from all classes share the same covariance matrix.

Linear Discriminant Analysis (LDA) assumes that the data from each class follows a multivariate Gaussian distribution and tries to minimize the total probability of misclassification~\cite{johnson2014applied}. Furthermore, the covariance matrices of all classes are assumed to be equal. 
The classification rule in LDA is to classify $\mathbf{x}$ to the $g^{th}$ class if 
$
	\mathcal{L}_g(\mathbf{x} ) = \max \{\mathcal{L}_1(\mathbf{x}),\mathcal{L}_2(\mathbf{x} ),...,\mathcal{L}_G(\mathbf{x}) \}.
$
Here,
\begin{equation*}
	\mathcal{L}_g(\mathbf{x}) = \pi_g - \frac{1}{2}(\mathbf{x}-\boldsymbol{\mu}^{(g)})^T\boldsymbol{\Sigma}^{-1}(\mathbf{x}-\boldsymbol{\mu}^{(g)}), \text{ for } g =1,2,\dots,G,
\end{equation*}
 where $\pi_g = \log \frac{n_g}{n}$ is the logarithm of prior probability. Here, $\mathbf{A}^T$ is denoted as the transpose of matrix $\mathbf{A}$.\\
Another form of the discriminant function for a class $g$ is
\begin{equation*}
  \delta_g(\mathbf{x}) = (\boldsymbol{\mu}^{(g)})^T \boldsymbol{\Sigma}^{-1}\mathbf{x} - \frac{1}{2} (\boldsymbol{\mu}^{(g)})^T \boldsymbol{\Sigma}^{-1} \boldsymbol{\mu}^{(g)} + \pi_g.
\end{equation*}
The decision boundary between two classes $g$ and $h$ is where their corresponding discriminant functions are equal, i.e., 
$
   \delta_g(\mathbf{x}) = \delta_h(\mathbf{x}).
$
Simplifying this equation, we get
\begin{align*}
    \scalebox{1}{$
    (\boldsymbol{\mu}^{(g)})^T \boldsymbol{\Sigma}^{-1}\mathbf{x} - \frac{1}{2} (\boldsymbol{\mu}^{(g)})^T \boldsymbol{\Sigma}^{-1} \boldsymbol{\mu}^{(g)} + \pi_g
     = (\boldsymbol{\mu}^{(h)})^T \boldsymbol{\Sigma}^{-1}\mathbf{x} - \frac{1}{2} (\boldsymbol{\mu}^{(h)})^T \boldsymbol{\Sigma}^{-1} \boldsymbol{\mu}^{(h)} + \pi_h.$}
\end{align*}
This implies to
\begin{align*}
\scalebox{1}{$
(\boldsymbol{\mu}^{(g)} - \boldsymbol{\mu}^{(h)})^T\boldsymbol{\Sigma}^{-1}\mathbf{x} ++ \frac{1}{2}\left((\boldsymbol{\mu}^{(h)})^T\boldsymbol{\Sigma}^{-1} \boldsymbol{\mu}^{(h)} - (\boldsymbol{\mu}^{(g)})^T\boldsymbol{\Sigma}^{-1} \boldsymbol{\mu}^{(g)}\right) + \log\frac{n_g}{n_h} = 0.$}
\end{align*}
This equation represents a linear decision boundary.
The weight vector $\mathbf{a}$ that defines the orientation of the decision boundary is given by
$
\mathbf{a}_{gh} =\boldsymbol{\Sigma}^{-1} (\boldsymbol{\mu}^{(g)} - \boldsymbol{\mu}^{(h)}).
$

The intercept term $a_o$ that defines the position of the decision boundary is 
$
a_o = \frac{1}{2}\left((\boldsymbol{\mu}^{(h)})^T\boldsymbol{\Sigma}^{-1} \boldsymbol{\mu}^{(h)} - (\boldsymbol{\mu}^{(g)})^T\boldsymbol{\Sigma}^{-1} \boldsymbol{\mu}^{(g)}\right) + \log\frac{n_g}{n_h}.
$

Putting it all together, the decision boundary can be expressed in the linear form
$
    \mathbf{a}_{gh}^T \mathbf{x} + a_0 = 0.
$
This linear equation defines the hyperplane that separates the two classes in the feature space.

\vspace{-1mm}
\section{Methodologies}\label{sec:methodologies}
In practice, missing values can be presented in a training or test set. When only the training set contains missing values, one can follow the same strategy as in~\cite{NGUYEN2022108082,nguyen2021epem}, i.e., estimating the parameters by using a direct parameter estimation method and then using that for LDA classification. However, this strategy can only be used when there is no missing value in the test set. When missing values are present in the testing data, one could use imputation to address this, but that is a two-step process that could introduce noises and biases into the dataset. 

These issues motivate us to introduce WLDA, a technique for LDA classification under missing data without the need for imputation, to tackle this problem. Here, the basic idea is that 
when a feature value is missing, its contribution to the classification task should be zero. 
Therefore, we propose using a weight matrix so that when a feature has a high missing rate, its contribution should be small. Therefore, we employ the weighted missing matrix to penalize features with a missing rate $r$ with a penalty function $\rho(r)$.
By using the weighted missing matrix, the algorithm adjusts the influence of each feature based on its reliability, thus maintaining the integrity of the classification process.
\vspace{-1mm}

\subsection{Weighted missing Linear Discriminant Analysis}\label{sec:LDA_missing}
In this section, we first give some notations and then provide more detail about our proposal. We denote $\mathbf{m}_\mathbf{x}=(m_1,m_2,\dots,m_p)^T$ as a mask of an observation $\mathbf{x} =(x_1,x_2,\dots,x_p)^T \in \mathbb{R}_p$, i.e., 
\begin{equation}
    m_{i} =  \begin{cases}
    1, & \text{ if } x_i \text{ is observed}  \\
    0, & \text{ if } x_i \text{ is missing}
    \end{cases}, \text{ for } i=1,2,\dots,p.
\end{equation}
Now, assuming each feature $\boldsymbol{f}_i$ of $\mathbf{X}$ has missing values with the rate $r_i$, for $i=1,2,\dots,p$. 
If the missing rate is high, we want the value of the LDA function to be small. In other words, $\rho(r)$ should be an increasing function w.r.t missing rate $r$. Therefore, we propose using the following function, which is an inverse of observed rates, i.e., 
\begin{equation}
    \rho(r) = \frac{1}{1-r}, \text{ where } r \in [0;1).
\end{equation}
Now, we define a \textbf{weighted missing vector} by $\mathbf{w}=(w_1,w_2,\dots,w_p)^T$ where
\begin{equation*}
    w_i = \rho(r_i) = \frac{1}{1-r_i}, \text{ for } i=1,2,\dots,p.
\end{equation*}
Then, \textbf{weighted missing matrix} $\mathbf{W}_{\mathbf{x}}$ is defined as a diagonal matrix of vector $\mathbf{m}_{\mathbf{x}} \otimes \mathbf{w}$, where $\mathbf{m}_{\mathbf{x}} \otimes \mathbf{w}$ is the direct product of $\mathbf{m}_{\mathbf{x}}$ and $\mathbf{w}$, i.e., 
\begin{equation*}
    \mathbf{W}_{\mathbf{x}} = \text{diag}(m_1w_1,m_2w_2,\dots,m_pw_p)
\end{equation*}
Therefore, our proposed classification function is
\begin{equation}
    \mathcal{L}^{\mathbf{W}}_{g}(\mathbf{x}) = \pi_g - \frac{1}{2}(\mathbf{x} -\boldsymbol{\mu}^{(g)})^T(\mathbf{W}_{\mathbf{x}}\boldsymbol{\Sigma}^{-1}\mathbf{W}_{\mathbf{x}})(\mathbf{x}-\boldsymbol{\mu}^{(g)}),
\end{equation}
and the corresponding WLDA algorithm is presented in the algorithm~\ref{alg:WLDA}.

\begin{algorithm}
\caption{\textbf{WLDA}}\label{alg:WLDA}
    \hspace*{\algorithmicindent} \textbf{Input:} A dataset including training set $(\textbf{X}_{train}|\textbf{y}_{train})$ which has $n_{train}$ samples and test set $(\textbf{X}_{test}|\textbf{y}_{test})$ which has $n_{test}$ samples, and $\textbf{y}_{train}$ has $G$ classes. \\
    \hspace*{\algorithmicindent} \textbf{Output:}
    the predicted $\widehat{\textbf{y}}_{test}=(\widehat{\textbf{y}}_{test}^{i})_{i=1}^{n_{test}}$.
\begin{algorithmic}[1]
    \State $\widehat{\boldsymbol{\mu}}^{(g)},\widehat{\boldsymbol{\Sigma}} \leftarrow$ estimated covariance matrix by DPER algorithm. 
    \State $\pi_g \leftarrow$ the prior probability of class $g$ that is computed by the quotient $n_{g}$ and $n_{train}$, where $n_{g}$ is the number of elements which is label $g$ in $\textbf{y}_{train}$.
    \State $\mathbf{w} \leftarrow$ the weighted missing vector.
    \For {$\mathbf{x}_i \in \mathbf{X}_{test}$}
            \State $\mathbf{W}_{\mathbf{x}_i} \leftarrow$ the weighted missing matrix of $\mathbf{x}_i$.
            \State $\mathcal{L}^{\mathbf{W}}_{g}(\mathbf{x}_i) \leftarrow \pi_g - \frac{1}{2}(\mathbf{x}_{i} -\widehat{\boldsymbol{\mu}}^{(g)})^T(\mathbf{W}_{\mathbf{x}_i}\widehat{\boldsymbol{\Sigma}}^{-1}\mathbf{W}_{\mathbf{x}_i})(\mathbf{x}_{i}-\widehat{\boldsymbol{\mu}}^{(g)})$.
            \State
            $\widehat{\textbf{y}}_{test}^{i} \leftarrow  \argmax_{g\in L}(\mathcal{L}^{\mathbf{W}}_g(\mathbf{x}_i))$, where $L=\{1,2,\dots,G\}$.
    \EndFor
\end{algorithmic}
\end{algorithm}

\subsection{The WLDA algorithm}\label{description}

The WLDA algorithm is detailed in Algorithm \ref{alg:WLDA}. In step 1, the algorithm starts by estimating the class means ($\widehat{\boldsymbol{\mu}}^{(g)}$) and the covariance matrix ($\widehat{\boldsymbol{\Sigma}}$) using the DPER algorithm. This step is crucial as it forms the basis for the WLDA classification function. In step 2, the prior probability ($\pi_g$) for each class $g$ is computed by dividing the number of training samples in class $g$ ($n_g$) by the total number of training samples ($n_{train}$). This helps in incorporating the relative frequency of each class into the classification function. Next, the weighted missing vector ($\mathbf{w}$) is initialized, which will be used to account for the reliability of observed features in the dataset. This vector plays a crucial role in adjusting the influence of each feature based on the presence of missing values. After that, the algorithm iterates over each test sample $\mathbf{x}_i$ in the test set. For each sample, first, the weighted missing matrix ($\mathbf{W}_{\mathbf{x}_i}$) is constructed based on the missing values in $\mathbf{x}_i$. Then, the WLDA classification score ($\mathcal{L}^{\mathbf{W}}_{g}(\mathbf{x}_i)$) is computed for each class $g$. This score incorporates the prior probability and a quadratic form that adjusts for missing data. Lastly, the class label $\widehat{\textbf{y}}_{test}^{i}$ is assigned to the class $g$ that maximizes the WLDA classification score.

\textbf{\textit{Remarks.}} We have some following remarks for this algorithm: 
\begin{itemize}
    \item The weighted missing matrix depends on the missing rate of both training and test set, thus the decision boundaries of WLDA are also.
    \item If there are no missing values in both the training and test sets, WLDA is actually the same as LDA.
    \item In case missing values are only presented in the training set and the test set is fully observed, the weighted missing matrix $\mathbf{W}_{\mathbf{x}}$ is the same for all test samples $\mathbf{x}$. One importance is that WLDA utilizes different decision boundaries compared to LDA. The specific results can be found in Section~\ref{sec:theoretical_analysis}.
    \item When the training set is fully observed but the test data contains missing values, if any test sample is fully observed, WLDA and LDA use the same decision boundaries. However, when a test sample contains missing values, WLDA would fit it with the corresponding decision boundaries, and therefore it outshines LDA when dealing with missing values.
    \item Lastly, if missing values are presented in both training and test sets then WLDA shows outstanding performances because this model uses different decision boundaries and also adjusts these boundaries suitable to each test sample.
\end{itemize}

Second, the key distinction between WLDA and LDA lies in their utilization of different decision boundaries coupled with the integration of a weighted missing matrix. In addition, this is a diagonal matrix, thereby not significantly increasing the computational burden. Essentially, this retains the computational efficiency of traditional LDA, making it suitable for large datasets and ensuring that WLDA remains practical for real-world applications.



\subsection{Theoretical analysis}\label{sec:theoretical_analysis}
We start a session by presenting the formulas of WLDA's decision boundaries and then provide explicit formulas for the expectation, variance, and bias of the WLDA classification function. Hence, enhancing our understanding of the WLDA model's behavior. This is also vital for diagnosing model performance and making informed decisions. 
Now, recall the WLDA classification function
\begin{equation*}
    \mathcal{L}^{\mathbf{W}}_g(\mathbf{x}) = \pi_g - \frac{1}{2}(\mathbf{x} -\boldsymbol{\mu}^{(g)})^T(\mathbf{W}_{\mathbf{x}}\boldsymbol{\Sigma}^{-1}\mathbf{W}_{\mathbf{x}})(\mathbf{x}-\boldsymbol{\mu}^{(g)}).
\end{equation*}
\begin{theorem}\label{theorem boundaries}
    The decision boundary between class $g$ and class $h$ of WLDA is given
   \begin{equation}\label{WLDA boundaries}
\mathbf{u}_{gh}^T\mathbf{x} + u_o = 0,
\end{equation}
where \begin{equation*}
\mathbf{u}_{gh} =\mathbf{W}_{\mathbf{x}}\boldsymbol{\Sigma}^{-1}\mathbf{W}_{\mathbf{x}}(\boldsymbol{\mu}^{(g)} - \boldsymbol{\mu}^{(h)}),
\end{equation*}
and
\begin{align*}
\scalebox{1}{$
u_o = \frac{1}{2}\left((\boldsymbol{\mu}^{(h)})^T\mathbf{W}_{\mathbf{x}}\boldsymbol{\Sigma}^{-1}\mathbf{W}_{\mathbf{x}} \boldsymbol{\mu}^{(h)} - (\boldsymbol{\mu}^{(g)})^T\mathbf{W}_{\mathbf{x}}\boldsymbol{\Sigma}^{-1}\mathbf{W}_{\mathbf{x}} \boldsymbol{\mu}^{(g)}\right)
+ \log\frac{n_g}{n_h}.$}
\end{align*}
\end{theorem}

The proof of this theorem is provided in~\ref{proof boundaries}. This theorem shows that by adjusting the matrix $\mathbf{W}_{\mathbf{x}}$ in $\mathbf{W}_{\mathbf{x}}\boldsymbol{\Sigma}^{-1}\mathbf{W}_{\mathbf{x}}$, these boundaries can make WLDA deal with missing data in many scenarios. Now, we present the following theorem in order to show the properties of the WLDA function, whose proof can be found in~\ref{proof}.
\begin{theorem}\label{main theorem}
    Assume we have a data $\mathbf{X}$ is size of $n \times p$ which has $G$ classes and each class data $\mathbf{X}^{(g)} \sim \mathcal{N}(\boldsymbol{\mu}^{(g)},\boldsymbol{\Sigma})$ for $g=1,2,\dots,G$. For each instance $\mathbf{x}$, the expectation and variance of the WLDA classification function of $\mathbf{x}$ belongs to $g^{th}$ class given by
    \begin{align}
    \mathbb{E}[\mathcal{L}^{\mathbf{W}}_g(\mathbf{x})] = \pi_g - \frac{1}{2}\sum_{i=1}^p m_i^2 w_i^2,\\
    \mathbf{Var}(\mathcal{L}^{\mathbf{W}}_g(\mathbf{x})) = \frac{1}{2}\sum_{i=1}^p m_i^4 w_i^4,
    \end{align}
    where $\mathbf{m}=(m_1,m_2,\dots,m_p)$ and $\mathbf{w}=(w_1,w_2,\dots,w_p)$ are the mask and weighted missing vector of $\mathbf{x}$ respectively.
    Moreover, the bias of this function is
    \begin{equation}
        \mathbf{Bias}(\mathcal{L}^{\mathbf{W}}_g(\mathbf{x})) = \frac{1}{2}\left(p-\sum_{i=1}^p m_i^2 w_i^2\right).
    \end{equation}
\end{theorem}

This theorem outlines the statistical properties (expectation, variance, and bias) of the WLDA classification function, emphasizing the crucial role of the weighted missing matrix in determining these properties. The expected value of the classification score provides insight into its central tendency under the WLDA model, illustrating how the model adjusts based on the presence of missing values, as captured by the mask $\mathbf{m}$ and weighted missing vectors $\mathbf{w}$. It is evident that if $m_i = 0$ (the feature value is missing) for $i=1,2,\dots, p$, the corresponding term $m_i^2w_i^2$ will be zero, and the feature does not contribute to the expectation. When feature value is present, the term $m_i^2w_i^2 = w_i^2 = \rho^2(r_i)$, contributes to the expectation. The overall expectation decreases as the increase of the high missing rate of present features, impacts the model's classification decision.

The variance quantifies the uncertainty or variability of the classification score, which is essential for assessing the reliability and stability of the WLDA model's classification decisions. Similar to the expectation, for $i=1,2,\dots,p$, feature $i$ contributes $\rho^4(r_i)$ to the variance if the corresponding feature is present with the missing rate of $r_i$. This means that a higher missing rate of present features leads to higher variance. A higher variance suggests more variability in the classification scores, indicating less confident decisions when features have higher weights and are present.

Bias measures the systematic error introduced by missing data, reflecting the difference between the expected classification score and the true classification score. The bias formula shows how the WLDA model adjusts its classification rule to compensate for missing data, providing a systematic approach to understanding and mitigating the impact of missing values on classification performance. The trade-off between variance and bias is also shown in these formulae, i.e., when feature $i$ is present with a missing rate of $r_i$, the variance increases $\rho^4(r_i)$ while the bias decreases $\rho^2(r_i)$, for $i=1,2,\dots,p$. Moreover, if there is no missing value, then the bias equals zero.

In summary, the theorem establishes a theoretical foundation for WLDA, ensuring that its properties are well-understood and theoretically justified, paving the way for further improvements and extensions of the WLDA method.

\section{Explanation strategies for the WLDA model}\label{sec:Explanation}
Interpreting the WLDA model, like many machine learning models, can be challenging. To enhance interpretability, we investigate three complementary strategies: correlation visualization, boundary similarity analyses, and Shapley values.
\vspace{-1mm}
\subsection {Correlation visualization}
We first examine the correlation through correlation heatmaps, local Squared Error correlation heatmaps, and subtraction correlation heatmaps as proposed in \cite{pham2024correlation}. The details of these plots are as follows

\begin{itemize}
    \item \textbf{\textit{Correlation heatmap}}: This heatmap visually represents the correlation matrix between features. It uses a color gradient to show the strength and direction of correlations, typically ranging from -1 (strong negative correlation) to 1 (strong positive correlation), with 0 indicating no correlation. To convert a covariance matrix $\boldsymbol{\Sigma}$ to a correlation matrix $\boldsymbol{R}$, we use the following formula
$
    \boldsymbol{R} = \boldsymbol{D}^{-1} \boldsymbol{\Sigma} \boldsymbol{D}^{-1}.    \label{correlation}
$
where $\boldsymbol{D}$ is a diagonal matrix containing the standard deviations of the features on the diagonal. 
\item  \textbf{\textit{Subtraction correlation heatmap}}: Also known as a difference heatmap, compares the ground truth correlation matrix $\boldsymbol{R}_{\text{true}}$ with an estimated correlation matrix $\boldsymbol{R}_{\text{est}}$. It visualizes where and to what extent the correlations between features differ between the two matrices
$
    \Delta{\boldsymbol{R}} = \boldsymbol{R}_{\text{true}} - \boldsymbol{R}_{\text{est}}. \label{SubEM}
$
The elements $\Delta{r_{ij}} = r_{ij}^{\text{true}} - r_{ij}^{\text{est}}$ of $\Delta{\boldsymbol{R}}$ represents the directional differences in correlation coefficients $r_{ij}$ between the ground truth and estimated matrices. This heatmap emphasizes whether these differences are positive or negative, providing a clear indication of the direction and magnitude of variation. 
\item \textbf{\textit{Local squared error correlation heatmap}}: The heatmap enhances the understanding of correlation estimation by visually depicting discrepancies between estimated correlation matrix $\boldsymbol{R}_{\text{est}}$ and true correlation matrix $\boldsymbol{R}_{\text{true}}$. The Squared Error is computed as
$
    \Delta \boldsymbol{R}_{\text{squared}} = (\Delta \boldsymbol{R}) \circ (\Delta \boldsymbol{R}). \label{SqrEM}
$
\end{itemize}

\subsection{Decision boundaries analysis}
Decision boundaries play a vital role in the interpretability of WLDA because these provide a visual representation of how different classes are separated in the feature space. This helps in understanding how well the algorithm distinguishes between different classes. Moreover, based on those boundaries, one can infer which features are most influential in distinguishing between classes. This insight is essential for interpretability, as it allows you to see the direct impact of each feature on the classification. In this analysis, we also focus on how the decision boundaries change with the change in missing rates.

\subsection{Shapley values}
Shapley values~\cite{lundberg2017unified} offer a systematic method to quantify the individual contribution of each feature towards model predictions. By calculating Shapley values, we can discern the exact impact of each feature value on the model's predictions, thereby improving our ability to interpret and trust the outcomes of WLDA and other comparable models. Moreover, we also use the mean absolute of Shapley values to illustrate the effects of each feature value on the classification results.

Examining these aspects not only elucidates the inner workings of the WLDA model but also enhances trust and transparency, crucial for applications in critical domains where interpretability is as important as accuracy.




\vspace{-1mm}
\section{Experiments}\label{sec:experiments}
\vspace{-1mm}
\subsection{Experiment settings}
In this work, we compare the performance of our proposed algorithm to the two-step procedures (imputation - parameter estimation) in the classification task. We experiment with four existing methods, from traditional techniques such as KNN Imputaion (KNNI)~\cite{zhang2012nearest} and Multiple Imputation by Chained Equation (MICE)~\cite{buuren2010mice} to progressive techniques such as SOFT-IMPUTE~\cite{mazumder2010spectral} and conditional Distribution-based Imputation of Missing Values with Regularization (DIMV)~\cite{vu2023conditional}. We implemented these methods using the \textit{fancyimpute} \footnote{https://github.com/iskandr/fancyimpute}, \textit{scikit-learn}~\cite{scikit-learn}, and \textit{DIMV Imputation} packages \footnote{\url{https://github.com/maianhpuco/DIMVImputation}}, with default configurations. Each experiment is repeated 10 times and run on an Intel Core i7 CPU with 4 cores, 8 threads, 2.11GHz base speed, and 16GB RAM. The codes for the experiments are available at \url{https://github.com/DangLeUyen/WLDA}.
\begin{table}[h]
    \caption{Description of datasets used in the experiments}
    \label{tab:dataset}
    \centering
    \begin{tabular}{@{\extracolsep\fill}cccc}
    \toprule
     \textbf{Datasets}  &  \textbf{\#Features} &\textbf{\#Samples} &\textbf{Classes distribution} \\
    \midrule
    Iris  &  4 & 150 & $(50,50,50)$ \\    
    Thyroid  &  5 & 215 & $(150,35,30)$ \\
    User  &  5 & 403 & $(50,129,122,130)$ \\
    \botrule
    \end{tabular}
\end{table}

The datasets for the experiments are sourced from the \textit{scikit-learn}~\cite{scikit-learn} (Iris dataset), and Machine Learning Database Repository at the University of California, Irvine~\footnote{http://archive.ics.uci.edu/ml} (Thyroid and User datasets). Table~\ref{tab:dataset} summarizes more details about datasets. Since any dataset can be rearranged so that missing columns are at the last and the samples are not classified if they are fully missing, we suppose that at least one (first) sample be fully observed. So the missing value just be at the last $(p - 1)$ features. Here, we simulate missing values in the remaining features by randomly generating the ratio between the number of missing entries and the total number of entries in the features with missing rates ranging from 15\% to 75\%, with a common spacing of 15\%. The performance of comparison algorithms is evaluated by using the accuracy score, which is the proportion of right predictions and the total number of predictions made. 
\subsection{Result and analysis}\label{sec:results}
In this section, we delve into the classification performance of various imputation methods applied to datasets with missing values. We provide a comprehensive analysis of accuracy scores and standard deviations across different missing rates to evaluate the robustness and reliability of each method. Additionally, we compare the covariance matrices of different methods using correlation heatmaps and explore the interpretability of WLDA results through the analysis of decision boundaries and Shapley values.
\subsubsection{Classification performances}

This subsection compares the proposed method with the other methods based on classification accuracy under two scenarios: when only the training data is missing and when both the training and test data are missing, as represented in Table~\ref{LDA-trainmiss} and Table~\ref{LDA-bothmiss}, respectively. Bold values represent the best results for each dataset.

\begin{table}[!htp]
\caption{The accuracy score (mean $\pm$ standard deviation) when missing values is in the training set only.} 
\label{LDA-trainmiss}
\centering
\addtolength{\tabcolsep}{-2.5pt}
\begin{tabular}{@{\extracolsep\fill}ccccccc}
\toprule
\multirow{2}{*}{\parbox{1cm}{\textbf{Dataset}}} & \multirow{2}{*}{\parbox{1cm}{\textbf{Missing rate}}} & \multicolumn{5}{c}{\textbf{Accuracy (mean $\pm$ standard deviation)}} \\
\cmidrule{3-7}
&& {\textbf{WLDA}} & \textbf{KNNI} & \textbf{MICE} & \textbf{Soft-Impute} & \textbf{DIMV} \\
\midrule
\multirow{5}{4em}{\textbf{Iris}} 
& 15\% & \textbf{1.000 $\pm$ 0.000} & 0.867 $\pm$ 0.021 & 0.860 $\pm$ 0.020 & 0.843 $\pm$ 0.052 & 0.860 $\pm$ 0.013 \\
& 30\% & \textbf{1.000 $\pm$ 0.000} & 0.857 $\pm$ 0.040 & 0.860 $\pm$ 0.020 & 0.727 $\pm$ 0.061 & 0.857 $\pm$ 0.015 \\
& 45\% & \textbf{0.987 $\pm$ 0.016} & 0.830 $\pm$ 0.050 & 0.843 $\pm$ 0.021 & 0.750 $\pm$ 0.067  & 0.860 $\pm$ 0.020 \\
& 60\% & \textbf{0.990 $\pm$ 0.015} & 0.807 $\pm$ 0.084 & 0.830 $\pm$ 0.041 & 0.667 $\pm$ 0.075 & 0.837 $\pm$ 0.038 \\
& 75\% & \textbf{0.987 $\pm$ 0.016} & 0.797 $\pm$ 0.066 & 0.803 $\pm$ 0.087 & 0.707 $\pm$ 0.084 & 0.843 $\pm$ 0.037 \\
\midrule
\multirow{5}{4em}{\textbf{Thyroid}} 
& 15\% & \textbf{0.933 $\pm$ 0.016} & 0.881 $\pm$ 0.013  & 0.881 $\pm$ 0.013 &  0.895 $\pm$ 0.016 & 0.884 $\pm$ 0.010 \\
& 30\% & \textbf{0.914 $\pm$ 0.011} & 0.886 $\pm$ 0.013 & 0.884 $\pm$ 0.015 & 0.907 $\pm$ 0.000 & 0.886 $\pm$ 0.013 \\
& 45\% & \textbf{0.937 $\pm$ 0.015} & 0.886 $\pm$ 0.022 & 0.893 $\pm$ 0.015 & 0.914 $\pm$ 0.015 & 0.895 $\pm$ 0.012 \\
& 60\% & \textbf{0.944 $\pm$ 0.021} & 0.893 $\pm$ 0.011 & 0.860 $\pm$ 0.048 & 0.933 $\pm$ 0.024 & 0.893 $\pm$ 0.021 \\
& 75\% & \textbf{0.933 $\pm$ 0.034} & 0.888 $\pm$ 0.014 & 0.826 $\pm$ 0.044 & 0.926 $\pm$ 0.017 & 0.884 $\pm$ 0.021  \\
\midrule
\multirow{5}{4em}{\textbf{User}} 
& 15\% & \textbf{0.937 $\pm$ 0.014} & 0.786 $\pm$ 0.018 & 0.794 $\pm$ 0.010 & 0.759 $\pm$ 0.032 & 0.791 $\pm$ 0.010 \\
& 30\% & \textbf{0.935 $\pm$ 0.020} &  0.760 $\pm$ 0.047 & 0.777 $\pm$ 0.033 & 0.684 $\pm$ 0.034 & 0.779 $\pm$ 0.034 \\
& 45\% & \textbf{0.936 $\pm$ 0.014} & 0.732 $\pm$ 0.057 & 0.748 $\pm$ 0.051 & 0.593 $\pm$ 0.030 & 0.756 $\pm$ 0.056 \\
& 60\% & \textbf{0.941 $\pm$ 0.020} & 0.741 $\pm$ 0.048 &	0.705 $\pm$ 0.066 &	0.444 $\pm$ 0.057 &	0.765 $\pm$ 0.043 \\
& 70\% & \textbf{0.911 $\pm$ 0.026} & 0.749 $\pm$ 0.055 &	0.642 $\pm$ 0.077 &	0.446 $\pm$ 0.075 &	0.769 $\pm$ 0.036 \\
\botrule
\end{tabular}
\end{table}
The experiment results in table~\ref{LDA-trainmiss} demonstrate that WLDA consistently outperforms other imputation methods across varying missing rates in the training set for the Iris, Thyroid, and User datasets. Specifically, WLDA achieves the highest accuracy with minimal standard deviation, maintaining near-perfect performance even at high missing rates. For instance, on the Iris dataset, WLDA maintains a high accuracy of $0.987 \pm 0.016$ even at a 75\% missing rate, showcasing its robustness. Conversely, Soft-Impute consistently shows the lowest performance, particularly as the missing rate increases. For instance, on the Iris dataset, Soft-Impute achieves accuracy scores of $0.843$ and $0.707$ at 15\% and 75\% missing rates, respectively, which are $0.157$ and $0.280$ lower than WLDA at the same rates, highlighting its limitations with datasets containing substantial missing data. KNNI, MICE, and DIMV show stable but moderate performance, typically ranging between $0.642$ and $0.890$ in accuracy. 
Overall, these findings underscore WLDA as the most reliable method, consistently delivering the highest accuracy with the least variability compared to KNNI, MICE, Soft-Impute, and DIMV.

\begin{table}[!htp]
\caption{The accuracy score (mean $\pm$ standard deviation) when missing values is in both training and test sets.}\label{LDA-bothmiss}
\centering
\addtolength{\tabcolsep}{-2.5pt}
\begin{tabular}{@{\extracolsep\fill}ccccccc}
\toprule
\multirow{2}{*}{\parbox{1cm}{\textbf{Dataset}}} & \multirow{2}{*}{\parbox{1cm}{\textbf{Missing rate}}} & \multicolumn{5}{c}{\textbf{Accuracy (mean $\pm$ standard deviation)}} \\
\cmidrule{3-7}
&& {\textbf{WLDA}} & \textbf{KNNI} & \textbf{MICE} & \textbf{Soft-Impute} & \textbf{DIMV} \\
\midrule
\multirow{5}{4em}{\textbf{Iris}} 
& 15\% & \textbf{0.977 $\pm$ 0.021} & 0.860 $\pm$ 0.047&	0.837 $\pm$ 0.046&	0.757 $\pm$ 0.075&	0.847 $\pm$ 0.037 \\
& 30\% & \textbf{0.970 $\pm$ 0.035} &  0.820 $\pm$ 0.062&	0.813 $\pm$ 0.037&	0.663 $\pm$ 0.126&	0.823 $\pm$ 0.040 \\
& 45\% & \textbf{0.947 $\pm$ 0.027} & 0.797 $\pm$ 0.064	&0.770 $\pm$ 0.050&	0.670 $\pm$ 0.059&	0.757 $\pm$ 0.056 \\
& 60\% & \textbf{0.923 $\pm$ 0.060} & 0.720 $\pm$ 0.058	&0.757 $\pm$ 0.045&	0.643 $\pm$ 0.070&	0.757 $\pm$ 0.054 \\
& 75\% & \textbf{0.917 $\pm$ 0.060} & 0.697 $\pm$ 0.048	&0.743 $\pm$ 0.070&	0.653 $\pm$ 0.083&	0.740 $\pm$ 0.074 \\
\midrule
\multirow{5}{4em}{\textbf{Thyroid}} 
& 15\% & \textbf{0.940 $\pm$ 0.021} &0.884 $\pm$ 0.028&	0.872 $\pm$ 0.021&	0.874 $\pm$ 0.030 &	0.872 $\pm$ 0.021 \\
& 30\% & \textbf{0.933 $\pm$ 0.016}&0.865 $\pm$ 0.033&0.863 $\pm$ 0.037&	0.865 $\pm$ 0.044&0.867 $\pm$ 0.036 \\
& 45\% & \textbf{0.923 $\pm$ 0.021} & 0.814 $\pm$ 0.040 & 0.833 $\pm$ 0.027 &	0.823 $\pm$ 0.032 &	0.833 $\pm$ 0.033 \\
& 60\% & \textbf{0.921 $\pm$ 0.035} &	0.842 $\pm$ 0.052 &	0.814 $\pm$ 0.021 &	0.833 $\pm$ 0.047 &	0.816 $\pm$ 0.016 \\
& 75\% & \textbf{0.907 $\pm$ 0.023} &	0.835 $\pm$ 0.064 &	0.805 $\pm$ 0.036 &	0.819 $\pm$ 0.046 &	0.821 $\pm$ 0.033 \\
\midrule
\multirow{5}{4em}{\textbf{User}} 
& 15\% & \textbf{0.832 $\pm$ 0.034} &	0.715 $\pm$ 0.032 &	0.712 $\pm$ 0.034 & 0.702 $\pm$ 0.030 &	0.710 $\pm$ 0.022 \\
& 30\% & \textbf{0.741 $\pm$ 0.064} &	0.643 $\pm$ 0.044 &	0.664 $\pm$ 0.036	& 0.583 $\pm$ 0.045	& 0.665 $\pm$ 0.037 \\
& 45\% & \textbf{0.693 $\pm$ 0.035} &	0.611 $\pm$ 0.061 &	0.607 $\pm$ 0.036 &	0.491 $\pm$ 0.054 &	0.616 $\pm$ 0.031 \\
& 60\% & \textbf{0.620 $\pm$ 0.047} &	0.560 $\pm$ 0.030 &	0.570 $\pm$ 0.044	& 0.448 $\pm$ 0.043	& 0.559 $\pm$ 0.037 \\
& 75\% & \textbf{0.569 $\pm$ 0.057} &	0.562 $\pm$ 0.042 &	0.512 $\pm$ 0.042 &	0.393 $\pm$ 0.042 &	0.507 $\pm$ 0.034 \\
\botrule
\end{tabular}
\end{table}

Table~\ref{LDA-bothmiss}, again, illustrates that the WLDA method achieves higher accuracy than other methods like KNNI, MICE, Soft-Impute, and DIMV when handling missing values in both training and test sets across all datasets. Several key conclusions can be drawn: Firstly, as the proportion of missing instances increases, the performance of all methods generally declines. For example, on the Iris dataset, WLDA's accuracy decreases from 0.977 with 15\% missing values to 0.923 with 60\% missing values, indicating a difference of 0.054. In contrast, KNNI experiences a more significant drop from 0.860 to 0.720 over the same range of missing values, with a gap of 0.140, which is 0.086 higher than WLDA's decline. Furthermore, WLDA demonstrates improved efficiency compared to other methods, particularly in scenarios with high levels of missing data. For example, on the Iris dataset with 75\% missing values, WLDA achieves an accuracy of 0.917, substantially outperforming KNNI (0.697), MICE (0.743), Soft-Impute (0.653), and DIMV (0.740).

In conclusion, the experimental results clearly establish WLDA as the optimal technique in both scenarios—when missing values are present exclusively in the training set and when they occur in both training and test sets—across the Iris, Thyroid, and User datasets. WLDA consistently achieves the highest accuracy with minimal standard deviation across all evaluated missing rates (15\% to 75\%), demonstrating significant superiority over KNNI, MICE, Soft-Impute, and DIMV. 

\subsubsection{Explainability analysis}\label{sec:interpretable_WLDA}
In this section, we examine the explainability of the methods under comparison on the Iris dataset in an end-to-end manner, starting with pre-modeling by analyzing the change in correlation matrices, and then interpretable modeling by investigating decision boundaries and Shapley values.



\begin{figure}[!htp]
    \centering
    \begin{subfigure}[b]{\columnwidth}
        \centering
        \includegraphics[width=0.9\textwidth]{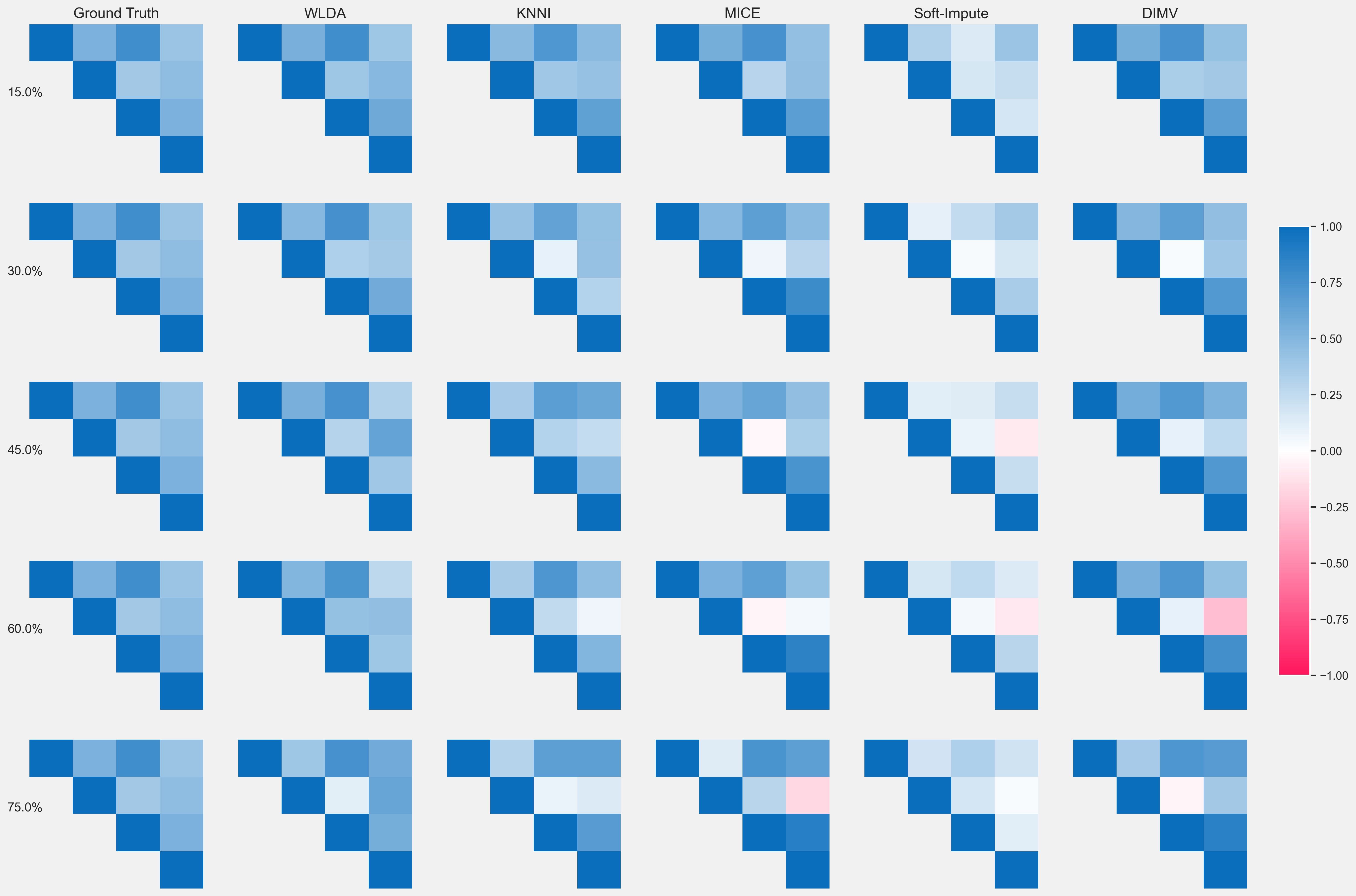}
        \caption{Correlation heatmap by using criteria~\eqref{correlation}}
        \label{fig:cor_mat}
    \end{subfigure}
    
    \begin{subfigure}[b]{0.49\columnwidth}
        \centering
        \includegraphics[width=0.85\textwidth]{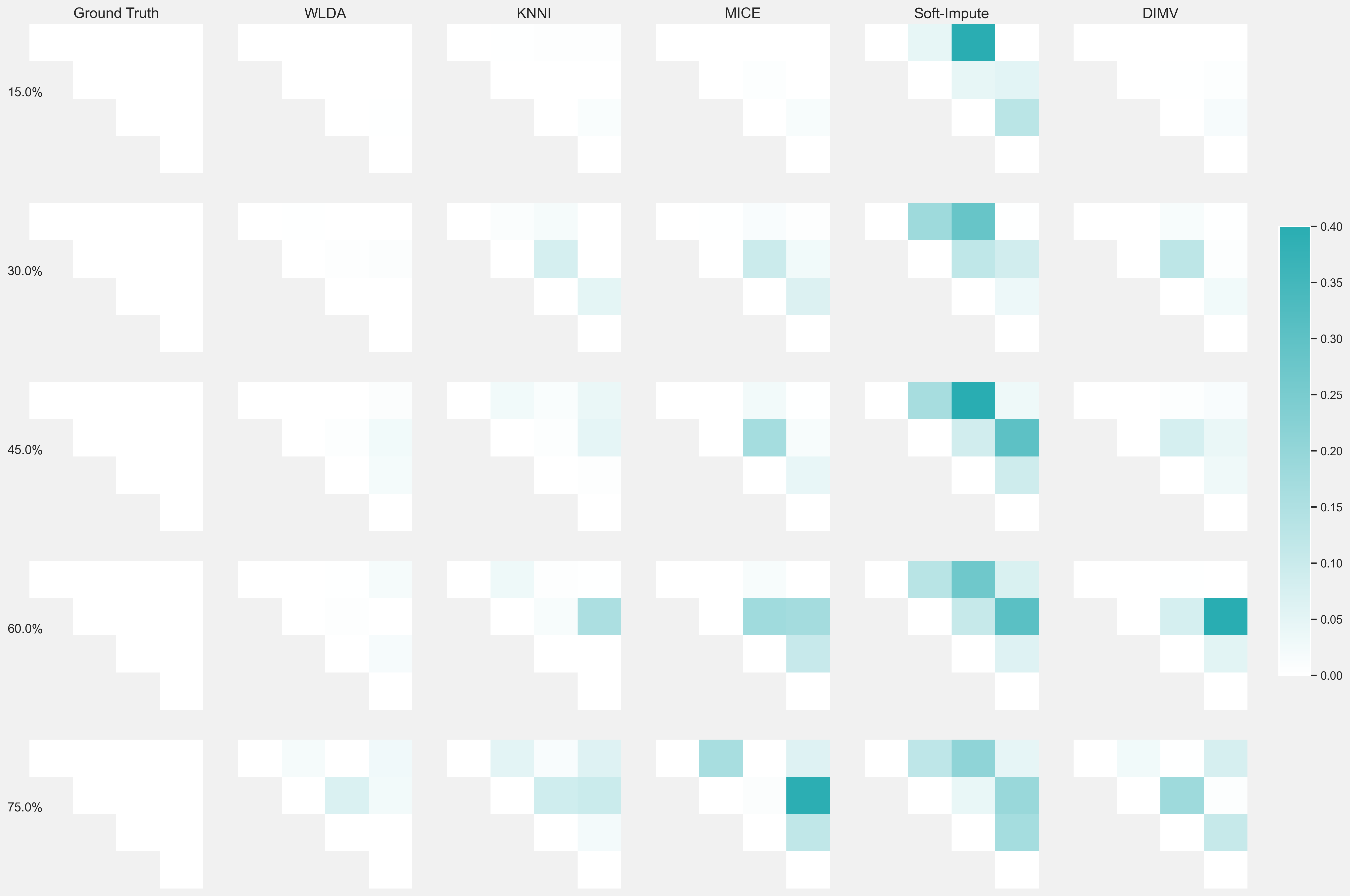}
        \caption{Local Squared Error correlation heatmap by using criteria~\eqref{SqrEM}}
        \label{fig:SqrEM}
    \end{subfigure}
    \hfill
    \begin{subfigure}[b]{0.49\columnwidth}
        \centering
        \includegraphics[width=0.85\textwidth]{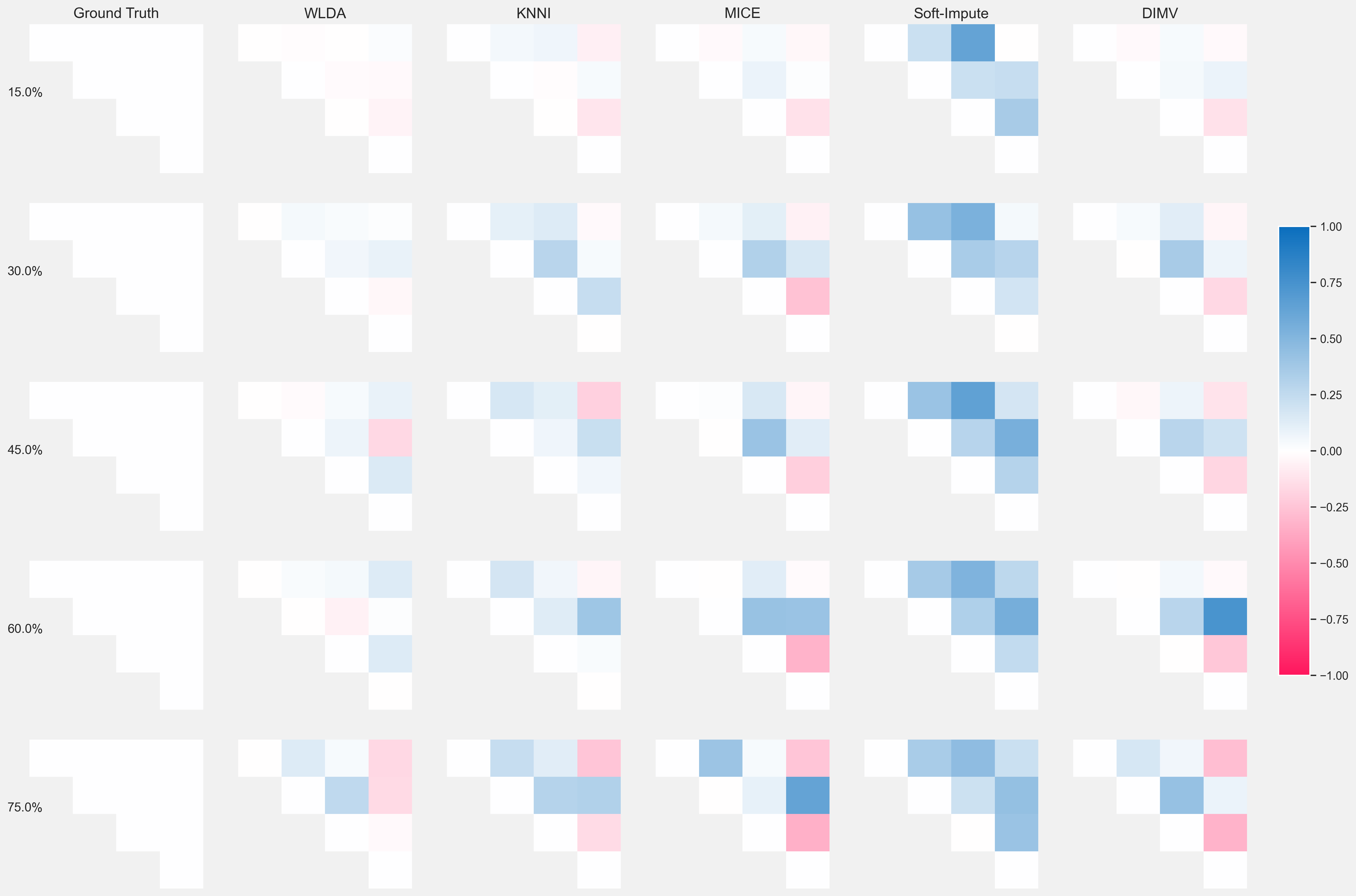}
        \caption{Subtraction correlation heatmap by using criteria~\eqref{SubEM}}
        \label{fig:SubEM}
    \end{subfigure}
    
    \caption{Correlation heatmap change of comparison methods on the Iris dataset with different missing rates}
    \label{fig:combined}
\end{figure}

First, in terms of correlation matrices, we illustrate the changes in the correlation matrices of the comparison methods at different missing rates in Figure~\ref{fig:combined}. Specifically, the correlation heatmap (Figure~\ref{fig:cor_mat}) provides a visual representation of the relationship between features under varying missing rates. The local squared error correlation heatmap (Figure~\ref{fig:SqrEM}) further enhances understanding by highlighting discrepancies between the estimated and true correlation matrices. Additionally, the subtraction correlation heatmap (Figure~\ref{fig:SubEM}) clearly shows the directional differences in correlation coefficients, providing clear insights into the direction and magnitude of these variations. 
Note that WLDA relies on DPER \cite{NGUYEN2022108082} for covariance matrix estimation. However, in the figures, to emphasize that fact, we label the corresponding column with ``WLDA" instead of DPER. Overall, WLDA, consistently preserves the true correlation structure across all missing rates, demonstrating superior robustness and suitability for datasets across missing rates. For example,  the subtraction correlation heatmap (Figure~\ref{fig:SubEM}) shows that at a 15\% missing rate, WLDA, DIMV closely match the Ground Truth with minimal deviations, while KNNI, MICE, and Soft-Impute show a little more discrepancies. As the missing rate increases to 30\% and 45\%, WLDA still approximates the Ground Truth closely, with slight deviations, whereas KNNI, MICE, and Soft-Impute exhibit increased deviations, especially in middle cells. DIMV starts to show more noticeable discrepancies. At 60\% and 75\% missing rates, WLDA, despite more pronounced deviations, remains the most accurate method. In contrast, KNNI, MICE, and Soft-Impute show significant deviations, and DIMV shows more substantial discrepancies. This parameter estimation precision partly explains the superior performance of WLDA during the classification phase.

\begin{table}[!h]
\caption{The coefficients of WLDA's decision boundaries across all missing rates on the Iris dataset, normalized by $w_0$.}
    \label{tab:normalized_coefficients}
    \centering
    \begin{tabular}{@{\extracolsep\fill}cccccc}
        \toprule
        \textbf{Missing rate} & \textbf{Boundary} & \textbf{Sepal length} & \textbf{Sepal width} & \textbf{Petal length} & \textbf{Petal width} \\
        \midrule
        \multirow{3}{*}{15\%} & (0, 1) & -0.116 & -0.025 & -0.096 & -0.028 \\
        & (0, 2) & -0.103 & -0.027 & -0.081 & -0.024 \\
        & (1, 2) & -0.090 & -0.030 & -0.065 & -0.021 \\
        \midrule
        \multirow{3}{*}{30\%} & (0, 1) & -0.121 & -0.028 & -0.085 & -0.023 \\
        & (0, 2) & -0.108 & -0.029 & -0.073 & -0.020 \\
        & (1, 2) & -0.094 & -0.030 & -0.061 & -0.016 \\
        \midrule
        \multirow{3}{*}{45\%} & (0, 1) & -0.125 & -0.036 & -0.068 & -0.020 \\
        & (0, 2) & -0.113 & -0.034 & -0.059 & -0.018 \\
        & (1, 2) & -0.101 & -0.032 & -0.051 & -0.016 \\
        \midrule
        \multirow{3}{*}{60\%} & (0, 1) & -0.129 & -0.029 & -0.068 & -0.017 \\
        & (0, 2) & -0.116 & -0.028 & -0.059 & -0.015 \\
        & (1, 2) & -0.103 & -0.028 & -0.050 & -0.013 \\
        \midrule
        \multirow{3}{*}{75\%} & (0, 1) & -0.134 & -0.023 & -0.065 & -0.012 \\
        & (0, 2) & -0.122 & -0.023 & -0.057 & -0.010 \\
        & (1, 2) & -0.109 & -0.023 & -0.048 & -0.009 \\
        \botrule
    \end{tabular}
\end{table}

Note that the decision boundary is a hyperplane. Therefore, to ensure a fair comparison of their values, we normalized the decision boundaries by dividing each coefficient by $u_o$ in~\eqref{WLDA boundaries}. In Table~\ref{tab:normalized_coefficients}, one can find that these coefficients show different trends. Across all boundaries when the missing rates increase, the coefficient of sepal length decreases, i.e., increasing negative impact. Moreover, this also consistently shows the highest negative value in a range of $[-0.134;-0.090]$, indicating its strong influence on the model. This is followed by petal length impacting significantly, ranging from $-0.096$ to $-0.048$, but with an opposite trend. Although sepal width and petal width have a relatively lesser impact, they also experience two different trends. While the coefficient of sepal width varies slightly at around $-0.03$, this value of the petal width declines from about $-0.024$ to less than half its peak missing rate. In summary, the sepal length is the most influential feature, with petal length also being highly significant, while sepal width and petal width have moderate impacts.

\begin{figure}[ht]
    \centering
    \includegraphics[width=0.9\textwidth]{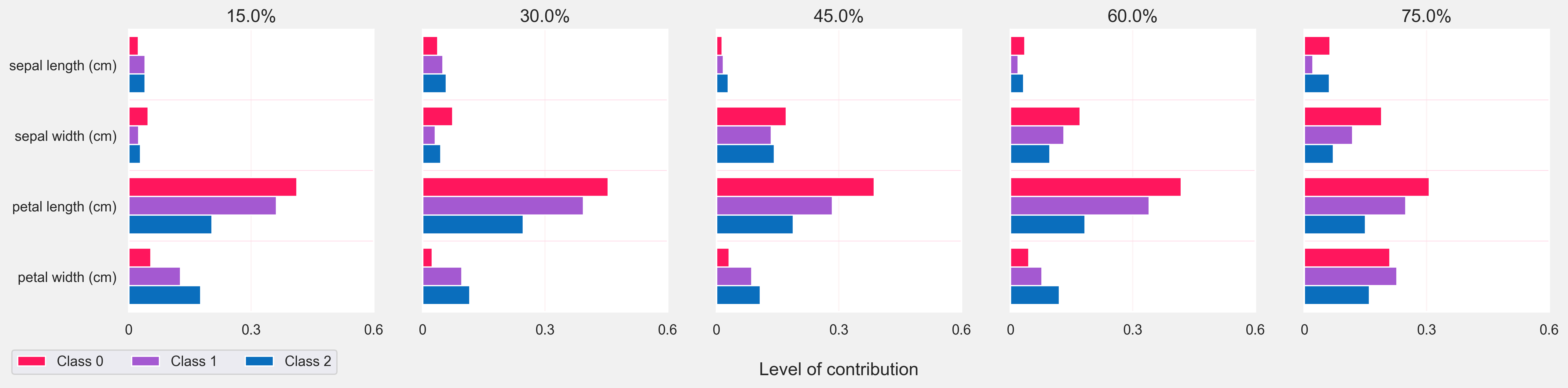}
    \caption{Global interpretability with feature importance bar plots of WLDA model by using mean absolute Shapley values.}
    \label{fig:shapley_general}
\end{figure}

\begin{figure}[!htp]
    \centering
    \includegraphics[width=0.9\textwidth]{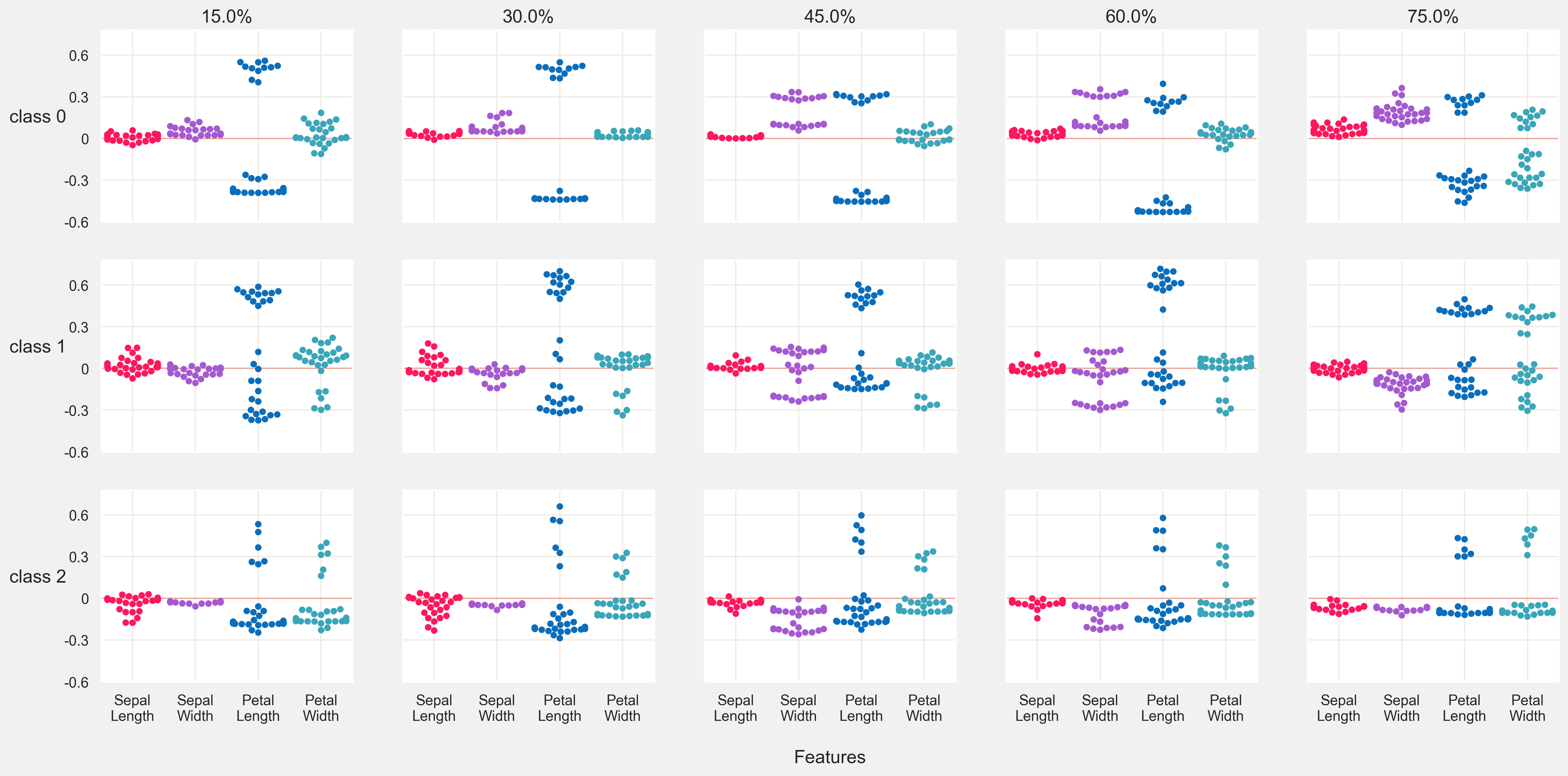}
    \caption{Level of the contribution of each feature to each class of WLDA model by using Shapley values.}
    \label{fig:Shapley values}
\end{figure}

Figure~\ref{fig:shapley_general} quantifies the contribution of each feature to the classification of WLDA varying missing rates. An interesting point is that although sepal length was identified as the feature with the highest coefficient impact through boundary analysis, petal length emerges as the most contributing feature to the classification results for each class across different missing rates. Moreover, at the highest missing rate (75\%), there is a more balanced contribution of feature values. Specifically, petal width shows a significant increase, followed by sepal width. Sepal length, however, continues to maintain a very low level of contribution. More insights into the positive or negative effects can be seen in Figure~\ref{fig:Shapley values}.

\section{Conclusion and future works}\label{sec:conclusion}
In this paper, we introduced Weighted missing Linear Discriminant Analysis (WLDA), a novel approach to Linear Discriminant Analysis (LDA) designed to effectively handle missing data without the need for imputation. 
Our method not only retains the interpretability of LDA but also provides comprehensive explanations throughout the modeling process using tools such as covariance matrices, decision boundaries, and Shapley values. Experimental results confirm that WLDA outperforms conventional methods, particularly in datasets with missing values in both training and test phases, demonstrating its robustness and superior accuracy. This work bridges the gap between explainable AI and practical application in critical fields, where transparency and reliability are paramount.

In addition, there are several areas for improvement in this investigation. Firstly, we only considered missing values in continuous features. Exploring scenarios where the labels are also missing would be intriguing. Secondly, our algorithms could be enhanced by incorporating a feature selection model, allowing a trade-off between running time and accuracy for potentially better performance. Next, WLDA can be used for a broader range of classification problems across various domains such as finance, healthcare, and cybersecurity to evaluate its effectiveness with different types of incomplete datasets. Moreover, developing advanced techniques for dynamically adjusting the weights in the missing matrix based on data mechanisms could further enhance the model’s performance and adaptability. Finally, conducting a deeper theoretical analysis of WLDA’s properties in different scenarios, including its worst-case performance and behavior under various missing data patterns, will provide a more comprehensive understanding of the method.
\backmatter

\bibliography{sn-bibliography}

\begin{appendices}

\section{Proof for the theorem~\ref{theorem boundaries}}\label{proof boundaries}
\noindent \textbf{Theorem~\ref{theorem boundaries}}.
\textit{
The decision boundary between class $g$ and class $h$ of WLDA is given by
$\mathbf{u}_{gh}^T\mathbf{x} + u_o = 0,$
where $
\mathbf{u}_{gh} =\mathbf{W}_{\mathbf{x}}\boldsymbol{\Sigma}^{-1}\mathbf{W}_{\mathbf{x}}(\boldsymbol{\mu}^{(g)} - \boldsymbol{\mu}^{(h)}),
$
and
$
u_o = \frac{1}{2}\left((\boldsymbol{\mu}^{(h)})^T\mathbf{W}_{\mathbf{x}}\boldsymbol{\Sigma}^{-1}\mathbf{W}_{\mathbf{x}} \boldsymbol{\mu}^{(h)} - (\boldsymbol{\mu}^{(g)})^T\mathbf{W}_{\mathbf{x}}\boldsymbol{\Sigma}^{-1}\mathbf{W}_{\mathbf{x}} \boldsymbol{\mu}^{(g)}\right) 
+ \log\frac{n_g}{n_h}.
$
}
\begin{proof}
    By setting $\mathbf{P}=\mathbf{W}_{\mathbf{x}}\boldsymbol{\Sigma}^{-1}\mathbf{W}_{\mathbf{x}}$, the WLDA function of class $g$ becomes
    \begin{equation*}
    \mathcal{L}^{\mathbf{W}}_g(\mathbf{x}) = \pi_g - \frac{1}{2}(\mathbf{x} -\boldsymbol{\mu}^{(g)})^T\mathbf{P}(\mathbf{x}-\boldsymbol{\mu}^{(g)}).
\end{equation*}
Now, the boundary between class $g$ and class $h$ is
\begin{equation}\label{Boundarygh}
    \pi_g - \frac{1}{2}(\mathbf{x} -\boldsymbol{\mu}^{(g)})^T\mathbf{P}(\mathbf{x}-\boldsymbol{\mu}^{(g)})   = \pi_h - \frac{1}{2}(\mathbf{x} -\boldsymbol{\mu}^{(h)})^T\mathbf{P}(\mathbf{x}-\boldsymbol{\mu}^{(h)}).
\end{equation}
Notably, if $\mathbf{A}^T\mathbf{B}$ is a $1\times 1$ matrix then $\mathbf{A}^T\mathbf{B} = \mathbf{B}^T\mathbf{A}$ for any matrices $\mathbf{A},\mathbf{B}$ of suitable dimensions. Therefore, expending each side of equation~\eqref{Boundarygh} and reducing the term $\mathbf{x}^T\mathbf{P}\mathbf{x}$ we have
\begin{align*}
\scalebox{1}{$
    (\boldsymbol{\mu}^{(g)})^T \boldsymbol{P}\mathbf{x} - \frac{1}{2} (\boldsymbol{\mu}^{(g)})^T \boldsymbol{P} \boldsymbol{\mu}^{(g)} + \pi_g  
    = (\boldsymbol{\mu}^{(h)})^T \boldsymbol{P}\mathbf{x} - \frac{1}{2} (\boldsymbol{\mu}^{(h)})^T \boldsymbol{P} \boldsymbol{\mu}^{(h)} + \pi_h.$}
\end{align*}
Then
\begin{align*}
\scalebox{1}{$
(\boldsymbol{\mu}^{(g)} - \boldsymbol{\mu}^{(h)})^T\boldsymbol{P}\mathbf{x}+ \frac{1}{2}\left((\boldsymbol{\mu}^{(h)})^T\boldsymbol{P} \boldsymbol{\mu}^{(h)} - (\boldsymbol{\mu}^{(g)})^T\boldsymbol{P} \boldsymbol{\mu}^{(g)}\right) + \log\frac{n_g}{n_h} = 0. $}
\end{align*}
It is worth noting that $\mathbf{W}_\mathbf{x}$ and $\boldsymbol{\Sigma}^{-1}$ are symmetric matrices, thus $\mathbf{P}^T = (\mathbf{W}_{\mathbf{x}}\boldsymbol{\Sigma}^{-1}\mathbf{W}_{\mathbf{x}})^T = (\mathbf{W}_{\mathbf{x}})^T(\boldsymbol{\Sigma}^{-1})^T(\mathbf{W}_{\mathbf{x}})^T = \mathbf{W}_{\mathbf{x}}\boldsymbol{\Sigma}^{-1}\mathbf{W}_{\mathbf{x}} = \mathbf{P}$.\\
Now, put \begin{align*}
\mathbf{u}_{gh} &=\mathbf{P}(\boldsymbol{\mu}^{(g)} - \boldsymbol{\mu}^{(h)}),\\
u_o &= \frac{1}{2}\left((\boldsymbol{\mu}^{(h)})^T\boldsymbol{P} \boldsymbol{\mu}^{(h)} - (\boldsymbol{\mu}^{(g)})^T\boldsymbol{P} \boldsymbol{\mu}^{(g)}\right) + \log\frac{n_g}{n_h}.
\end{align*}
Therefore, the decision boundary between class $g$ and class $g$ is simplified to $\mathbf{u}_{gh}^T\mathbf{x} + u_o = 0.$
\end{proof}

\section{Proof for theorem~\ref{main theorem}}\label{proof}
\noindent \textbf{Theorem~\ref{main theorem}}.
\textit{
Assume we have a data $\mathbf{X}$ is size of $n \times p$ which has $G$ classes and each class data $\mathbf{X}^{(g)} \sim \mathcal{N}(\boldsymbol{\mu}^{(g)},\boldsymbol{\Sigma})$ for $g=1,2,\dots,G$. For each instance $\mathbf{x}$, the expectation and variance of the WLDA classification function of $\mathbf{x}$ belongs to $g^{th}$ class given by
$
    \mathbb{E}[\mathcal{L}^{\mathbf{W}}_g(\mathbf{x})] = \pi_g - \frac{1}{2}\sum_{i=1}^p m_i^2 w_i^2,
    \mathbf{Var}(\mathcal{L}^{\mathbf{W}}_g(\mathbf{x})) = \frac{1}{2}\sum_{i=1}^p m_i^4 w_i^4,
$
    where $\mathbf{m}=(m_1,m_2,\dots,m_p)$ and $\mathbf{w}=(w_1,w_2,\dots,w_p)$ are the mask and weighted missing vector of $\mathbf{x}$ respectively.
    Moreover, the bias of this function is
$
        \mathbf{Bias}(\mathcal{L}^{\mathbf{W}}_g(\mathbf{x})) = \frac{1}{2}\left(p-\sum_{i=1}^p m_i^2 w_i^2\right).
$
}

Before going into detail about the proof of this theorem, We should briefly summarize a lemma which is a corollary of \textit{theorem 1} in~\cite{bao2010expectation}.
\begin{lemma}\label{lemma}
    Let $\mathbf{x}$ be an $p \times 1$ random vector with mean $\boldsymbol{\mu}$ and covariance $\boldsymbol{\Sigma}$ and let $\mathbf{A}$ be a symmetric $p \times p$ matrix. Then, the expectation of the quadratic form $Q = \mathbf{x}^T\mathbf{A}\mathbf{x}$ is
    \begin{align}
        \mathbb{E}\left[Q\right] &= \boldsymbol{\mu}^T \mathbf{A} \boldsymbol{\mu} + \text{tr}(\mathbf{A} \boldsymbol{\Sigma}), \label{EQ}\\
        \textbf{Var}(Q) &= 4\boldsymbol{\mu}^T\mathbf{A}\boldsymbol{\Sigma}\textbf{A}\boldsymbol{\mu} + 2\textbf{Tr}(\mathbf{A}^2\boldsymbol{\Sigma}^2).\label{EQ^2}
    \end{align}
\end{lemma}
A poof of this lemma can be found in the proof of \textit{theorem 1} in~\cite{bao2010expectation}. Now we show the proof for \textbf{Theorem~\ref{main theorem}}
\begin{proof}
    We assume $\mathbf{x} \sim \mathcal{N}(\boldsymbol{\mu}^{(g)},\boldsymbol{\Sigma})$ provided that the sample $\mathbf{x}$ belongs to class $g^{th}$. Therefore, let $\mathbf{z} = x - \boldsymbol{\mu}^{(g)}$ then $\mathbf{z} \sim \mathcal{N}(\boldsymbol{0},\boldsymbol{\Sigma})$. By the properties of expectation
    \begin{align*}
        \mathbb{E}[\mathcal{L}^{\mathbf{W}}_g(\mathbf{x})] &= \mathbb{E}[\pi_g - \frac{1}{2}(\mathbf{x} -\boldsymbol{\mu}^{(g)})^T(\mathbf{W}_{\mathbf{x}}\boldsymbol{\Sigma}^{-1}\mathbf{W}_{\mathbf{x}})(\mathbf{x}-\boldsymbol{\mu}^{(g)})] \\
        &= \pi_g -\frac{1}{2}\mathbb{E}[\mathbf{z}^T(\mathbf{W}_{\mathbf{x}}\boldsymbol{\Sigma}^{-1}\mathbf{W}_{\mathbf{x}})\mathbf{z}].
    \end{align*}
    Note that $\boldsymbol{\mu}_\mathbf{z} = 0$ and $\mathbf{A} = \mathbf{W}_{\mathbf{x}}\boldsymbol{\Sigma}^{-1}\mathbf{W}_{\mathbf{x}}$ is a symmetric matrix, therefore we can apply the lemma~\ref{lemma} for $Q=\mathbf{z}^T(\mathbf{W}_{\mathbf{x}}\boldsymbol{\Sigma}^{-1}\mathbf{W}_{\mathbf{x}})\mathbf{z}$
    \begin{equation*}
        \mathbb{E}[Q] = \mathbf{Tr}(\mathbf{W}_{\mathbf{x}}\boldsymbol{\Sigma}^{-1}\mathbf{W}_{\mathbf{x}}\boldsymbol{\Sigma}).
    \end{equation*}
    Now, note that for any matrices $A, B$ of suitable dimensions, we have the trace property $\mathbf{Tr}(AB) = \mathbf{Tr}(BA)$. Therefore, 
    \begin{equation*}
        \mathbb{E}[Q] = \mathbf{Tr}(\mathbf{W}_{\mathbf{x}}\mathbf{W}_{\mathbf{x}}\boldsymbol{\Sigma}^{-1}\boldsymbol{\Sigma}) = \mathbf{Tr}(\mathbf{W}_{\mathbf{x}}^2).
    \end{equation*}
    Then, substitute $\mathbf{W}_{\mathbf{x}} = \text{diag}(m_1w_1,m_2w_2,\dots,m_pw_p)$ then
    \begin{equation}
         \mathbb{E}[Q] = \sum_{i=1}^p m_i^2w_i^2. \label{EQ = mw}
    \end{equation}
    Therefore, 
    \begin{equation*}
         \mathbb{E}[\mathcal{L}^{\mathbf{W}}_g(\mathbf{x})] = \pi_g - \frac{1}{2}\sum_{i=1}^p m_i^2 w_i^2.
    \end{equation*}
    We turn to the variance of $\mathcal{L}^{\mathbf{W}}_g(\mathbf{x})$. By the property of variance
    \begin{equation*}
        \mathbf{Var}(\mathcal{L}^{\mathbf{W}}_g(\mathbf{x})) = \frac{1}{4}\mathbf{Var}(Q).
    \end{equation*}
    Using the equation~\eqref{EQ^2} in lemma~\ref{lemma} with the notice of $\boldsymbol{\mu}_\mathbf{z} = 0$, we have
    \begin{align*}
    \mathbf{Var}(Q) &= 2\mathbf{Tr}((\mathbf{W}_{\mathbf{x}}\boldsymbol{\Sigma}^{-1}\mathbf{W}_{\mathbf{x}})^2\boldsymbol{\Sigma}^2) = 2\mathbf{Tr}(\mathbf{W}_{\mathbf{x}}^4) = 2\sum_{i=1}^p m_i^4 w_i^4.
    \end{align*}
    Therefore,
    \begin{equation*}
    \mathbf{Var}(\mathcal{L}^{\mathbf{W}}_g(\mathbf{x})) = \frac{1}{2}\sum_{i=1}^p m_i^4 w_i^4.
    \end{equation*}
    Finally, the bias of \(\mathcal{L}^{\mathbf{W}}_g(\mathbf{x})\) as an estimator of the true classification score is
    \begin{equation*}
        \text{Bias}(\mathcal{L}^{\mathbf{W}}_g(\mathbf{x})) = \mathbb{E}[\mathcal{L}^{\mathbf{W}}_g(\mathbf{x}) - \mathcal{L}_g(\mathbf{x})].
    \end{equation*}
    The true classification score assuming complete data is
    \begin{equation*}
        \mathcal{L}_g(\mathbf{x}) = \pi_g - \frac{1}{2}(\mathbf{x} - \boldsymbol{\mu}^{(g)})^T \boldsymbol{\Sigma}^{-1} (\mathbf{x} - \boldsymbol{\mu}^{(g)}).
    \end{equation*}
    From our earlier calculation, we found the equation~\eqref{EQ = mw} $\mathbb{E}[Q] = \sum_{i=1}^p m_i^2 w_i^2$. The main difference between the two terms is due to the weighting by \(\mathbf{W}_{\mathbf{x}}\). For complete data, \(\mathbf{W}_{\mathbf{x}}\) would be the identity matrix \(\mathbf{I}\), and hence:
    \begin{equation*}
        \sum_{i=1}^p m_i^2 w_i^2 = p, \quad \text{if there are no missing values}.
    \end{equation*}
    This implies that for complete data
$
        \mathbb{E}[\mathcal{L}_g(\mathbf{x})] = \pi_g - \frac{1}{2}p.
$
    Therefore,
$
     \textbf{Bias}(\mathcal{L}^{\mathbf{W}}_g(\mathbf{x})) = \left( \pi_g - \frac{1}{2} \sum_{i=1}^p m_i^2 w_i^2 \right) - \left( \pi_g - \frac{1}{2} p \right)  
      = \frac{1}{2}\left(p-\sum_{i=1}^p m_i^2 w_i^2\right).
$

\end{proof}
\end{appendices}

\end{document}